\documentclass[12pt]{article}
\usepackage{graphicx}
\usepackage{amsmath}
\usepackage[margin=1.0in]{geometry}
\usepackage{color, colortbl}
\usepackage{neuralnetwork}
\usepackage{hyperref}
\usepackage{float}
\usepackage[affil-it]{authblk}
\usepackage[english]{babel}
\definecolor{LightCyan}{rgb}{0.88,1,1}
\definecolor{LightRose}{rgb}{1,0.88,0.88}
\definecolor{LightGreen}{rgb}{0.88,1,0.88}

\title{Auto-encoders for Track Reconstruction in Drift Chambers for CLAS12}

\author[1]{Gagik Gavalian\thanks{Corresponding author}}
\affil[1]{Thomas Jefferson National Accelerator Facility, Newport News, VA 23606}
\date{}

\begin{document}

\begin{titlepage}
\maketitle
\begin{abstract}
In this article, we describe the development of machine learning models to assist the CLAS12 tracking algorithm
by identifying tracks through inferring missing segments in the drift chambers.  Autoencoders are used to reconstruct 
missing segments from track trajectories. Implemented neural network was able to reliably reconstruct missing 
segment positions with an accuracy of $\approx 0.35$ wires, and lead to the recovery of missing tracks with an accuracy of $>99.8\%$. 
\end{abstract}
\end{titlepage}

\section{Introduction}

\indent

The CLAS12\cite{Burkert:2020akg} detector is built around a six-coil toroidal magnet which divides the active detection into six azimuthal regions, called "sectors". The torus coils are approximately planar. Each sector subtends an azimuthal range of 60$^\circ$ from the mid-plane of one coil to the mid-plane of the adjacent coil. The “sector mid-plane” is an imaginary plane that bisects the sector’s azimuth. 
Charged particles in the CLAS12 detector are tracked using drift chambers\cite{Mestayer:2020saf} inside the toroidal magnetic field. There are six identical independent drift chamber systems in CLAS12 (one for each azimuthal sector). Each sector of drift chambers consists of three-chamber sets (called "regions"), and each region consists of two chambers called super-layers each of them containing 6 layers of wires perpendicular to particle trajectories in CLAS12. The tracks passing through drift chambers leave a signal in each of the layers (36 in total), which are broken down into 6 segments (one segment per super-layer). The tracking algorithm relies on forming track candidates on all combinations of segments (one from each super-layer).
With time some inefficiencies in the detector develop leading to missing segments in one (or more) of the super-layers, and this results in an efficiency drop in track identification. In this work, we investigate neural networks that can help improve the track finding efficiency when there are missing segments in some parts of the drift chambers.

\section{Reconstruction Procedure}

\indent

\begin{figure}[!ht]
\begin{center}
 \includegraphics[width=3.2in]{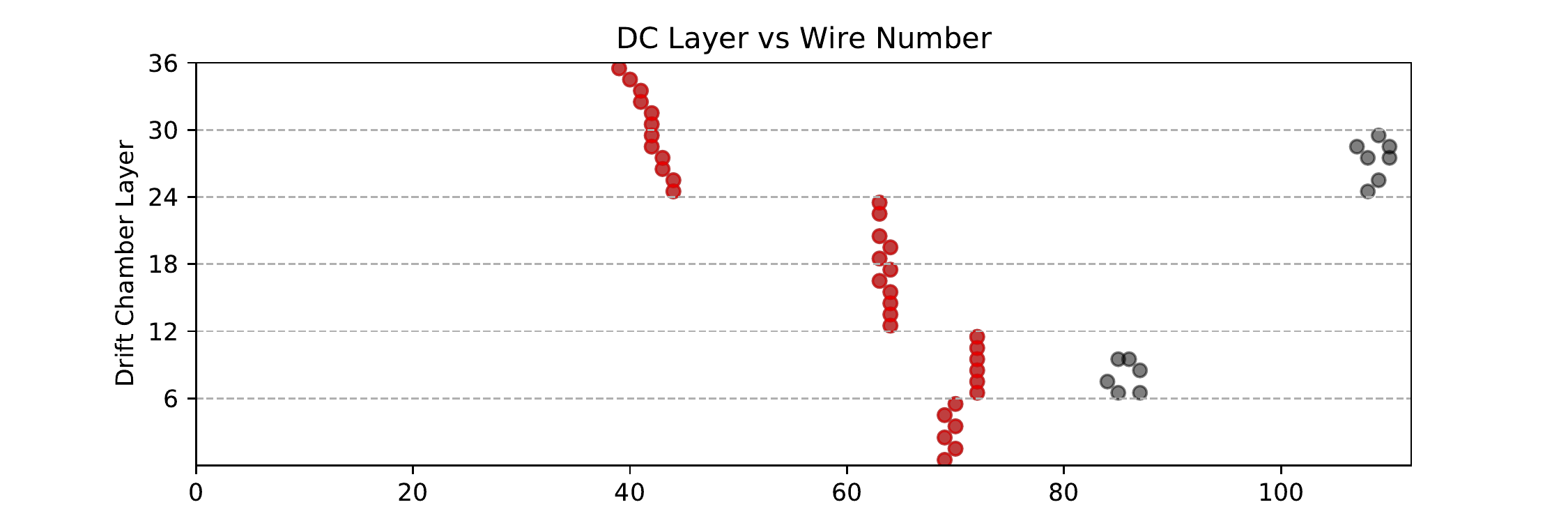}
 \includegraphics[width=3.2in]{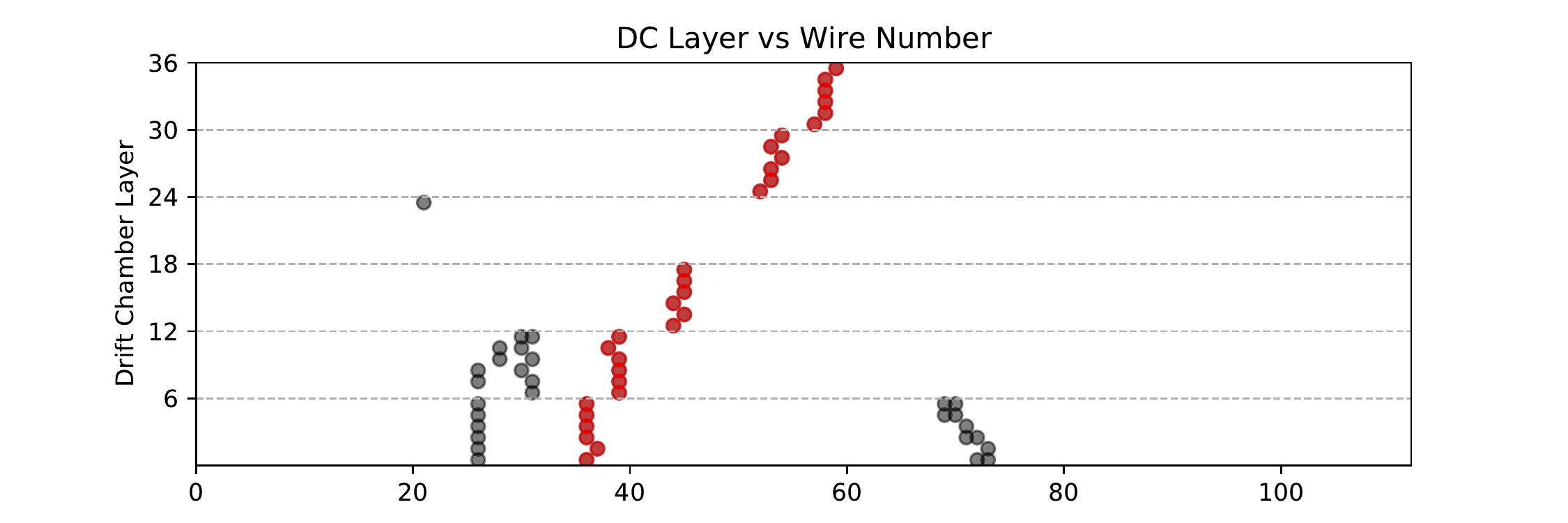}
  \includegraphics[width=3.2in]{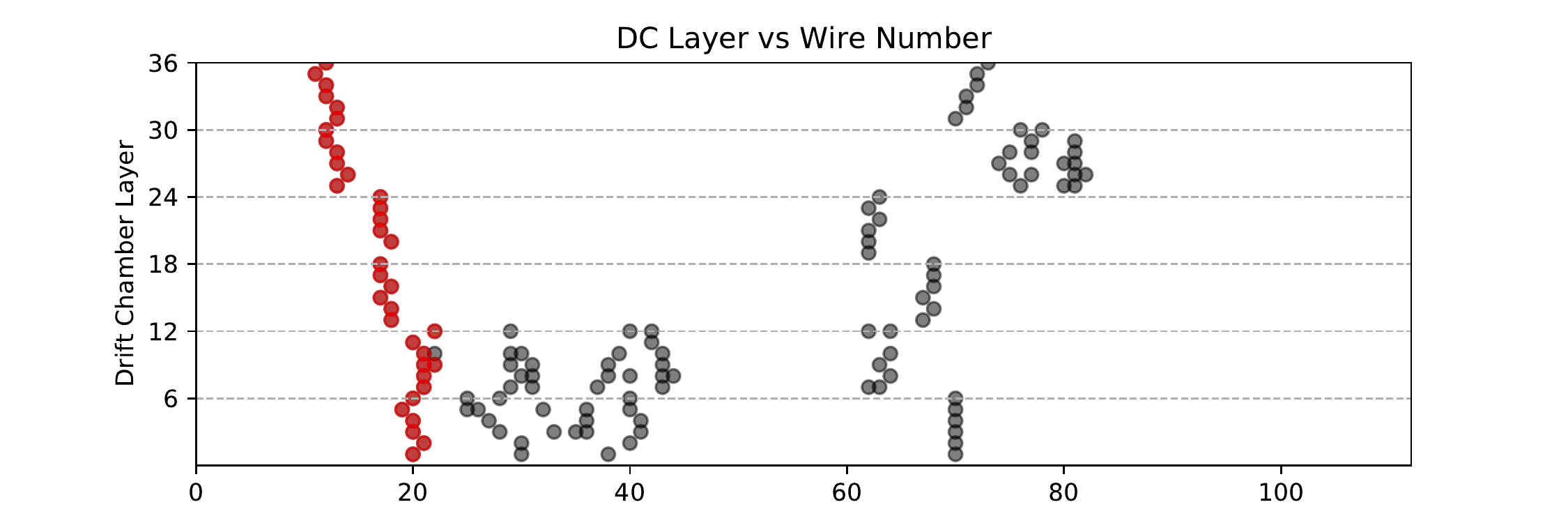}
   \includegraphics[width=3.2in]{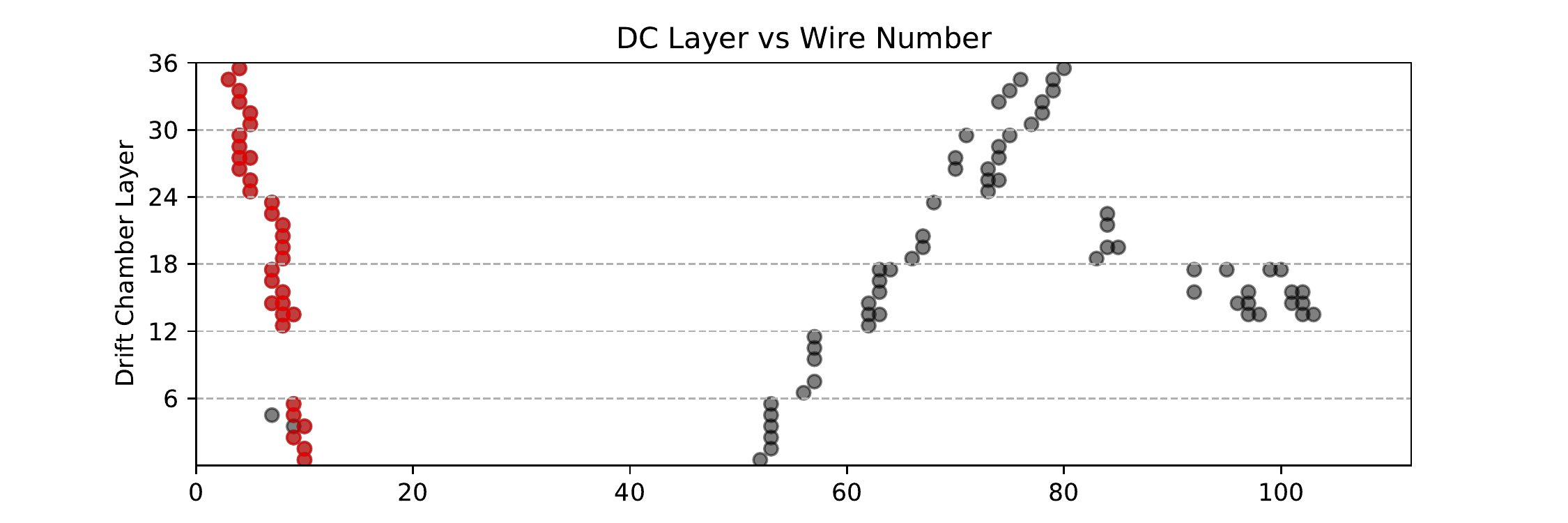}
\caption {Events from one of the sectors of drift chambers, where background hits (in gray) are plotted along with 
reconstructed track segments (red points). Events in the left column have all six segments present in the track, events
in the right column have one of the segments missing. The dashed lines on the plot show super-layer boundaries.}
 \label{dc:display}
 \end{center}
\end{figure}

A particle traveling through drift chambers leaves a signal in each of the 36 planes in the path of the particle. The signals in each super-layer are combined into segments. In Figure~\ref{dc:display} examples events are shown, where the horizontal lines on the plot correspond to the boundaries of each super-layer. The tracking algorithm constructs track candidates from composed segments by requiring one segment per super-layer for each track candidate. Each track candidate is fitted using Kalman-Filter to reconstruct a track, and depending on convergence it's either disregarded or marked as a good track for further considerations. This process of track candidate fitting is computationally intensive, and we already developed a neural network for CLAS12~\cite{Gavalian:2020oxg} detector which identifies correct combinations of segments to be considered by the tracking algorithm. The developed network provided accuracy of $>99.7\%$ and accelerated tracking code by a factor of 6, by composing a possible track candidate list using a neural network.

This procedure works well when all 6 super-layers have segments on the particle path. But with time drift chambers develop regions where the efficiency of producing a signal in the wire drops, and it is possible to end up with only 5 segments on the particle path, example of these events can be seen in Figure~\ref{dc:display} (right column). The current algorithm for track candidate identification classifier relies on 6 segment combinations to correctly identify tracks.

When missing segments are in the first or last super-layer series prediction can be successfully used to predict the last missing segment given 5 consecutive segments. We have already developed a series prediction neural network (using LSTMs)~\cite{Angelopoulos20P} capable of predicting the last missing segment. 

However, the missing segment can appear in any of the super-layers, and we need a reliable way of predicting the missing segment location, in order to pass complete track candidate information to our classifier network, which in turn can decide if the track is a valid track. To solve this problem we decided to use an auto-encoder type network.

\section{Network Architecture}

\indent

An auto-encoder is an unsupervised learning technique for neural networks that learns efficient data representations (encoding) by training the network to ignore signal “noise.” 
The auto-encoder network has three layers: the input, a hidden layer for encoding, and the output decoding layer. Using back-propagation, the unsupervised algorithm continuously trains itself by setting the target output values to equal the inputs. This forces the smaller hidden encoding layer to use dimensional reduction to eliminate noise and reconstruct the inputs.
Auto-encoder networks teach themselves how to compress data from the input layer into a shorter code, and then uncompress that code into whatever format best matches the original input. This process sometimes involves multiple autoencoders, such as stacked sparse auto-encoder layers used in image processing.

There are several types of auto-encoders:

{\bf De-noising auto-encoders:}  Using a partially corrupted input to learn how to recover the original undistorted input.
 More hidden encoding layers than inputs, and some use the outputs of the last auto-encoder as their input.

{\bf Contractive auto-encoder}: This uses an explicit “regularizer” that forces the model to learn a function that is robust against different variations of the input values.

\begin{figure}[!ht]
\begin{center}
 \includegraphics[width=3.0in]{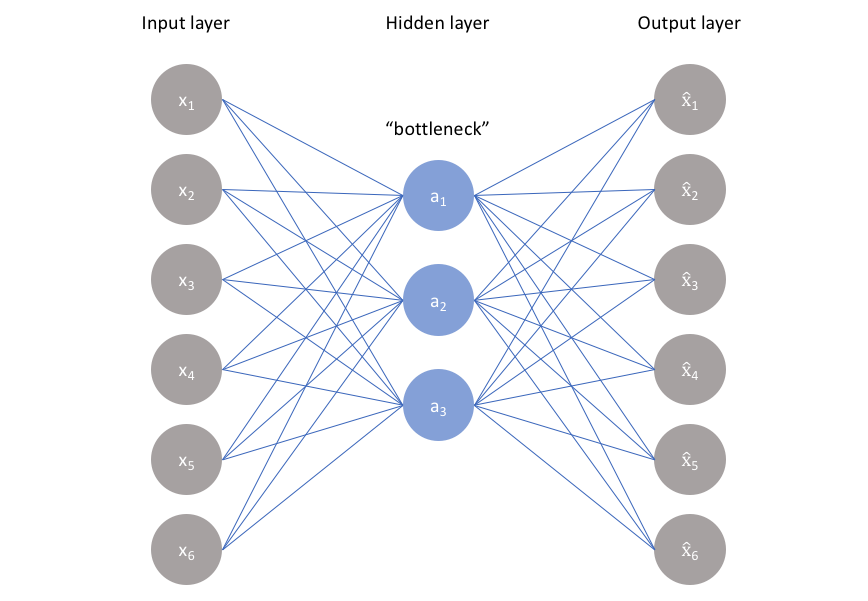}
 \includegraphics[width=3.0in]{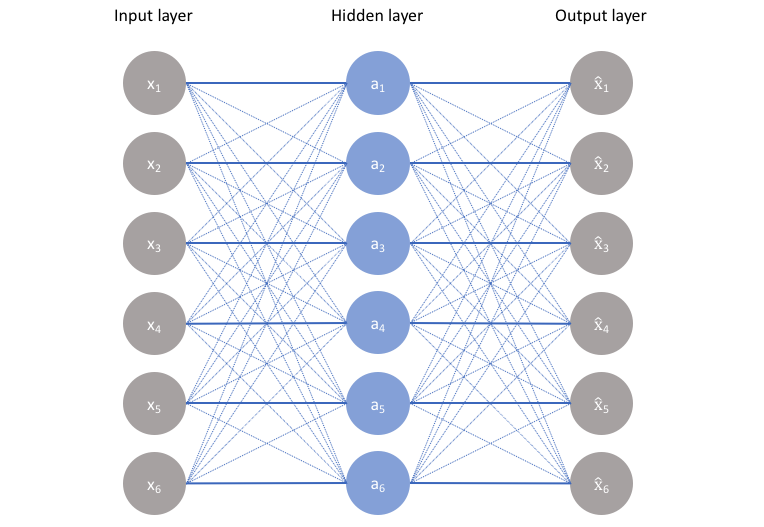}
\caption {Auto-encoder neural network architecture types.}
 \label{pic:autoencoders}
 \end{center}
\end{figure}

In Figure~\ref{pic:autoencoders}) different architectures of auto-encoder can be seen. For this study, we used 
an auto-encoder with the structure shown in Table~\ref{table:network}. Several different configurations were tried, and
we discovered this one to provide the best accuracy.
The network was implemented in Java using the machine learning library Neuroph~\cite{neuroph-2.98}. 

\begin{table}[!h]
\begin{center}
\begin{tabular}{|l|c|c|c|c|c|}
\hline
 & Input Layer & Hidden Layer 1 & Hidden Layer 2 & Hidden Layer 3 & Output Layer \\
\hline
\hline
Type & Dense Layer & Dense Layer &Dense Layer &Dense Layer &Dense Layer \\
\hline
Nodes & 6 & 12 & 6 & 12 & 6 \\
\hline
\end{tabular}
\end{center}
\caption{Neural Network architecture for resolving missing segments from track data.}
\label{table:network}
\end{table}

With this network, we plan to recover missing segment information. The input to the network is
the average wire position of each segment in for of a vector:

\begin{equation}
X = {x_1,x_2,x_3,x_4,x_5,x_6}
\end{equation}

There the index is the super-layer number for each segment, and the value is:

\begin{equation}
x_k = \sum_{i=1..N} \frac{w^k_i}{N}
\end{equation}

where N is number of hits in the segment, and $w^k_i$ is the wire position of the hit $i$
in super-layer $k$. And the output provided to the network is:

\begin{equation}
Y = (x_1,x_2,x_3,x_4,x_5,x_6)
\end{equation}

The corruption was introduced in the input set of training data by setting one of the values of the vector to "0",
and the neural network was trained to reconstruct the input data in the output without corruption. The procedure
of data corruption is discussed in the next section.

\section{Data Selection}
\label{section:data}

\indent

For this study, we selected events from processed data where a track was reconstructed in a given
sector and contained segments from all 6 super-layers. The input data set is a vector of 6 values,
representing the mean wire positions for each segment in a given super-layer:

\begin{equation}
 X = (x_1,x_2,x_3,x_4,x_5,x_6)
\end{equation}

and the output vector contains the same values as the input:

\begin{equation}
Y ={x_1,x_2,x_3,x_4,x_5,x_6}
\end{equation}

Two data sets were composed of this initial sample. One with one of the values of the input vector set to "0.0"
(chosen randomly):

\begin{equation}
  (x_1,x_2,x_3,x_4,x_5,x_6) \begin{cases}
    i = rndm(1..6)  & x_i = 0.0  \\
    X (x_1,0.0,x_3,x_4,x_5,x_6)   \text{$\rightarrow$}  & Y (x_1,x_2,x_3,x_4,x_5,x_6) \\
     \end{cases}
\end{equation}

In this given case i=2. For the second training set, each sample of X, and Y vectors
was extended to 6 samples, where in each sample one of the input vector values 
was set to "0.0":

\begin{equation}
  (x_1,x_2,x_3,x_4,x_5,x_6) \begin{cases}
   X (0.0,x_2,x_3,x_4,x_5,x_6)  \text{$\rightarrow$}  & Y (x_1,x_2,x_3,x_4,x_5,x_6)  \\
   X (x_1,0.0,x_3,x_4,x_5,x_6)   \text{$\rightarrow$}  & Y (x_1,x_2,x_3,x_4,x_5,x_6) \\
   X (x_1,x_2,0.0,x_4,x_5,x_6)  \text{$\rightarrow$}  &  Y (x_1,x_2,x_3,x_4,x_5,x_6) \\
   X (x_1,x_2,x_3,0.0,x_5,x_6)  \text{$\rightarrow$}  & Y (x_1,x_2,x_3,x_4,x_5,x_6) \\
   X (x_1,x_2,x_3,x_4,0.0,x_6)  \text{$\rightarrow$}  & Y (x_1,x_2,x_3,x_4,x_5,x_6) \\
   X (x_1,x_2,x_3,x_4,x_5,0.0)  \text{$\rightarrow$}  & Y (x_1,x_2,x_3,x_4,x_5,x_6)
  \end{cases}
\end{equation}

The second training sample is just ensuring that all combinations of missing segment information
is represented in the training sample. If a very large data set is used for training the first method of data
construction should be sufficient since many combinations of similar tracks with a random number of missing
segments randomly removed will be represented in the training set in abundance.

\section{Results}

\indent

For this study, we used tracks reconstructed with conventional algorithm and trained network using two data sets
(described in section \ref{section:data}). The training sample consisted of 5,000 samples, and the testing sample was 3,500 samples
(testing sample was different from the training sample). Two networks were trained and evaluated for this study. 
One with using random substitution in the input training set and one with creating 6 individual input and output pairs for each input data sample.
Then trained network was used to evaluate the testing data set. From the testing data set one of the nodes was set to "0.0", choosing the node 
randomly then the vector was provided as an input for the trained network and from the output, the value for that particular element of the vector 
was compared to the value of the same element in the input vector.

\begin{figure}[!ht]
\begin{center}
 \includegraphics[width=3.0in]{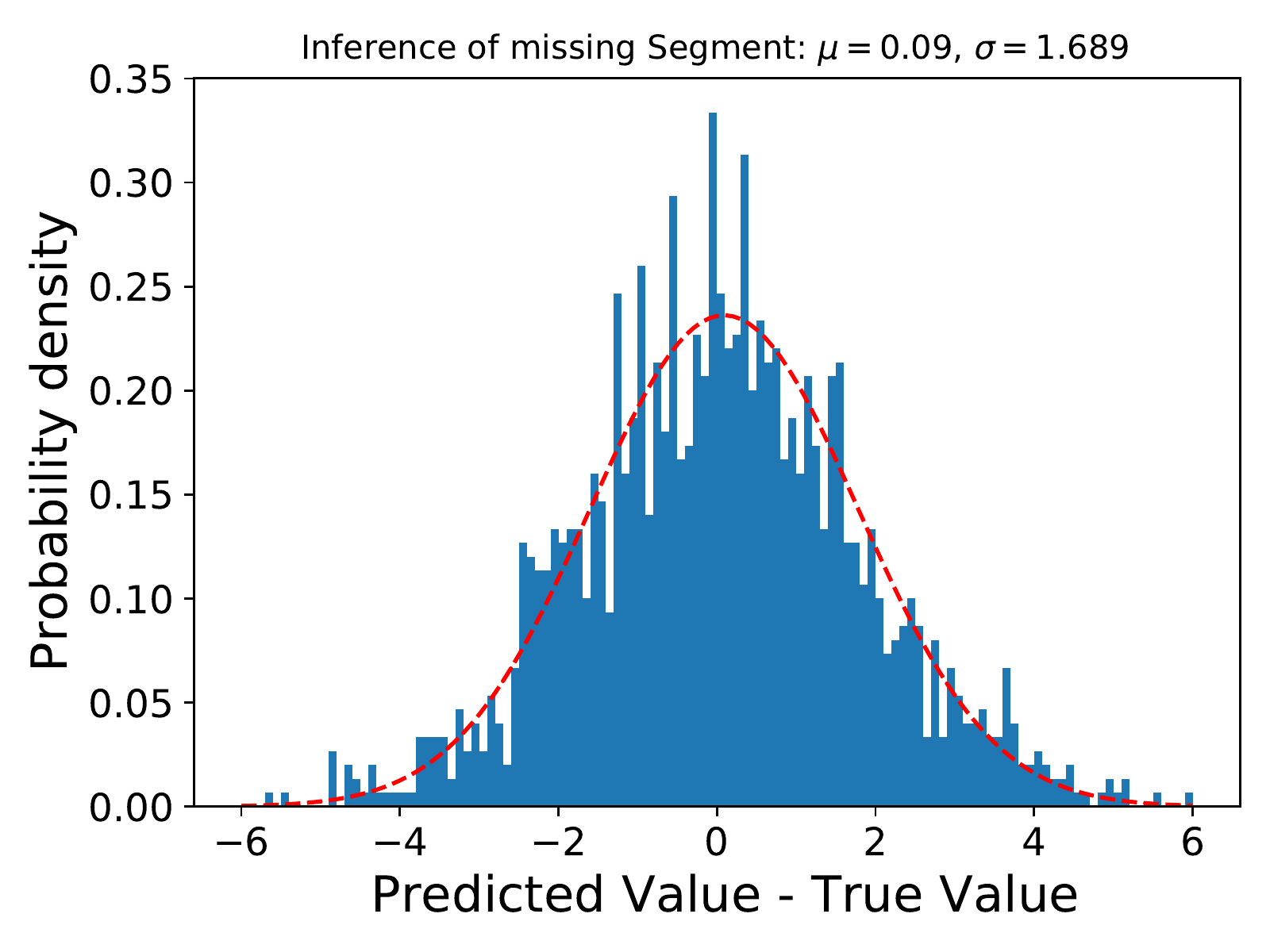}
 \includegraphics[width=3.0in]{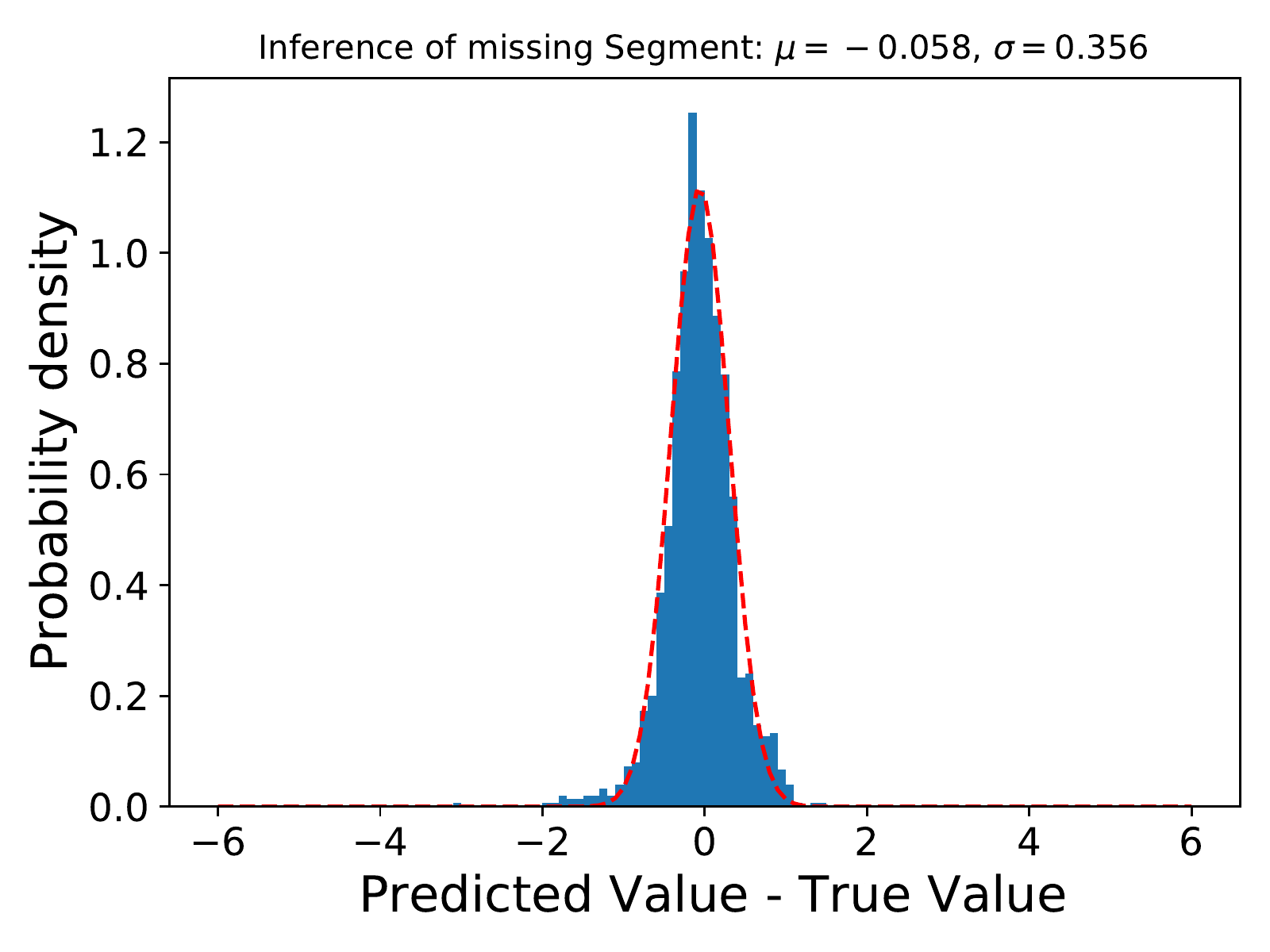}
\caption {Results of Neural Network Performance. The missing segment average wire numbers were inferred using trained network and the difference with plotted for regular data set (on the left) and extended data set (on the right). }
 \label{ml:results}
 \end{center}
\end{figure}

\begin{figure}[!ht]
\begin{center}
 \includegraphics[width=3.0in]{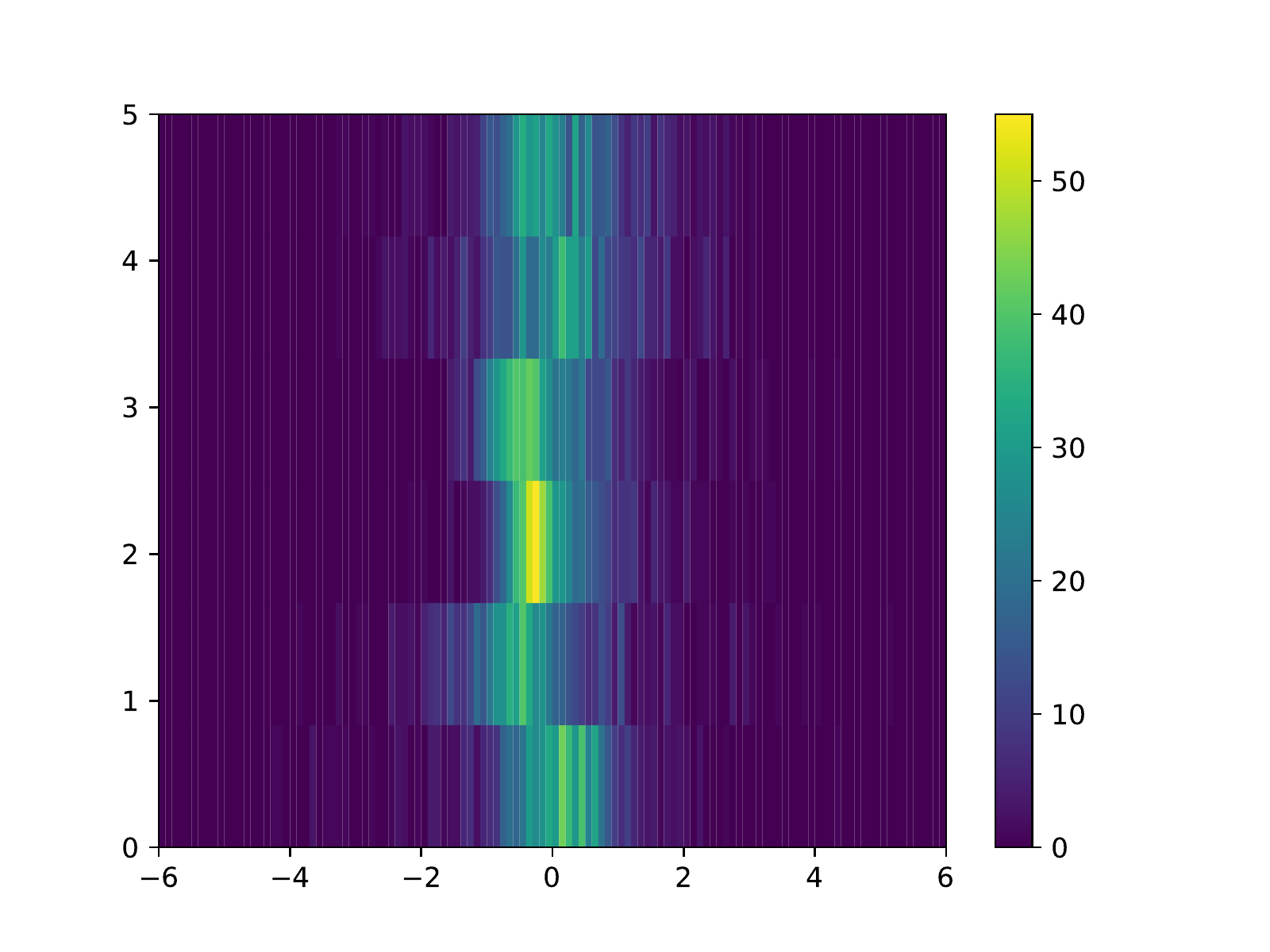}
 \includegraphics[width=3.0in]{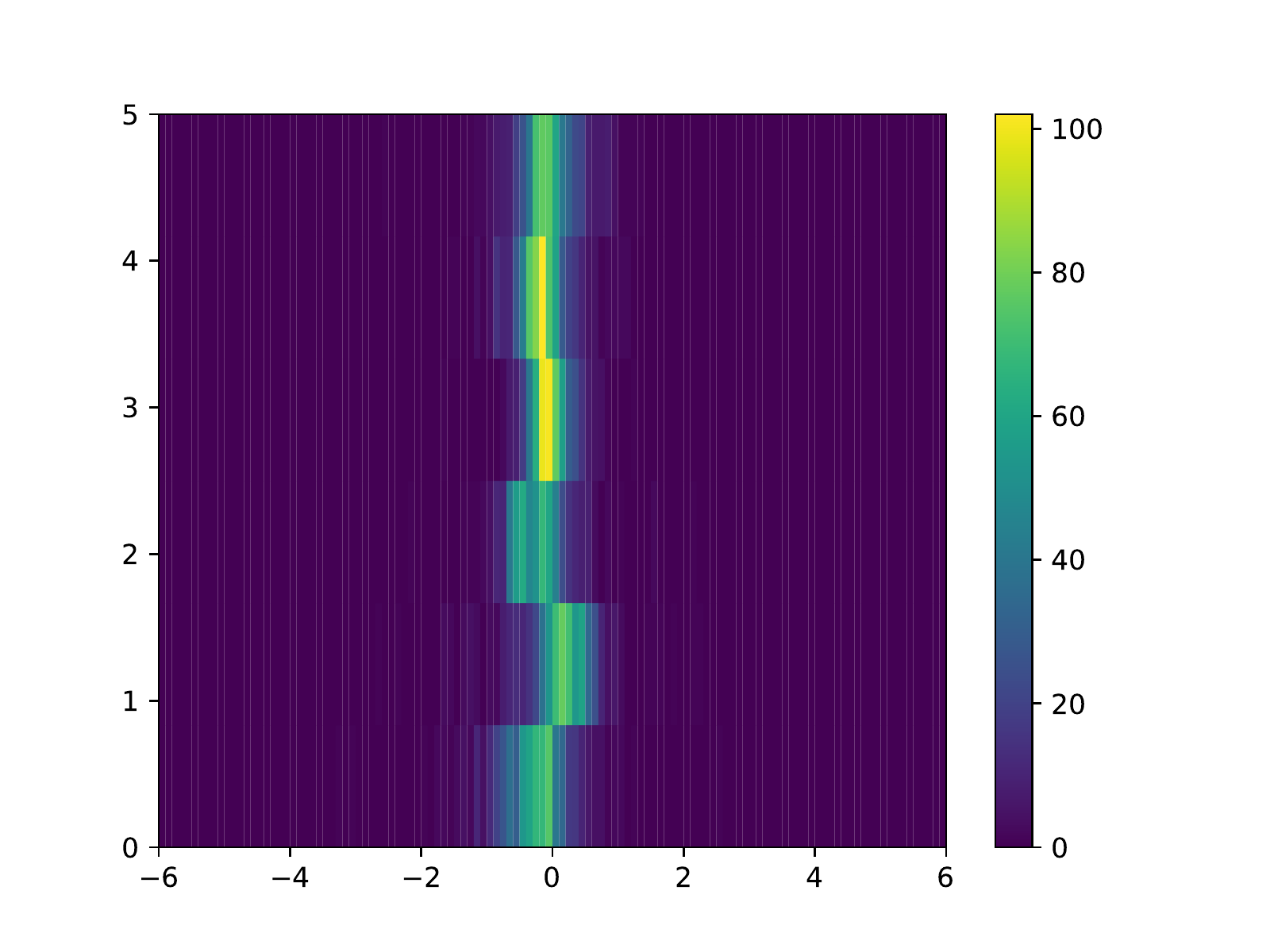}
\caption {Performance of neural network missing segment inference as a function of missing segment trained with regular data  set (on the left) and with extended data set (on the right).}
 \label{ml:results2d}
 \end{center}
\end{figure}

The results of network evaluation are shown in Figure~\ref{ml:results}, where the performance of both networks is presented with the corresponding fit. In Figure~\ref{ml:results} (left) the performance of the network trained with random "0.0" substitutions are presented, and the difference between the "True" segment value and inferred segment mean value is plotted. The average uncertainty of inference is $1.7$ wires. In Figure~\ref{ml:results} (right) the performance of the second network is shown with much better performance with the uncertainty of $0.36$ wires. In Figure~\ref{ml:results2d}  the performance of two networks is shown as a function of which corrupt node was reconstructed by the network. There are some systematic shifts depending on the knocked-out node, but they are well within the average error.

In theory with an increasing training sample, the performance of the first network should asymptotically reach the performance of the second network which has a more complete training sample, and the procedure of extending the training sample will not be necessary. We extended the training sample by duplication to overcome sparsity of data sample, and to illustrate that this technique works
with a smaller training sample.

\newpage
\bibliography{references}
\bibliographystyle{ieeetr}

\end{document}